\documentclass[12pt, letter]{article}

\usepackage[utf8]{inputenc}%
\usepackage[margin=1.2in,letterpaper,centering]{geometry}%
\usepackage{circuitikz}
\usepackage{subcaption}
\usepackage{todonotes}
\usepackage{graphicx}%
\usepackage{multirow}%
\usepackage{amsmath,amssymb,amsfonts}%
\usepackage{amsthm}%
\usepackage{mathrsfs}%
\usepackage{listings}
\usepackage{algorithmic}
\usepackage[title]{appendix}%
\usepackage{xcolor}%
\usepackage{textcomp}%
\usepackage{manyfoot}%
\usepackage{booktabs}%
\usepackage{algorithm}%
\usepackage{listings}%
\usepackage{pdflscape}%
\usepackage{adjustbox}%
\usepackage{cite}%

\usepackage{hyperref}%

\usepackage{listings}
\usepackage{xcolor}

\lstdefinestyle{pythonstyle}{
    language=Python,
    basicstyle=\ttfamily\small,
    commentstyle=\color{gray},
    keywordstyle=\color{blue},
    numberstyle=\tiny\color{gray},
    stringstyle=\color{red},
    breakatwhitespace=false,
    breaklines=true,
    captionpos=b,
    keepspaces=true,
    numbers=left,
    numbersep=5pt,
    showspaces=false,
    showstringspaces=false,
    showtabs=false,
    tabsize=2,
    frame=single,
    rulecolor=\color{black}
}

\begin{document}
\begin{center}
{\LARGE\bfseries 
The Optimiser Hidden in Plain Sight: Training with the Loss Landscape's Induced Metric}
\end{center}
\begin{center}
Thomas R. Harvey\footnote{\href{mailto:trharvey@mit.edu}{trharvey@mit.edu}}, 

\bigskip
{
	{\it NSF AI Institute for Fundamental Interactions, MIT, Cambridge, MA 02139, USA}\\[.5em]
}
\end{center}
\setcounter{footnote}{0} 
\bigskip\bigskip

\begin{abstract}
We present a class of novel optimisers for training neural networks that makes use of the Riemannian metric naturally induced when the loss landscape is embedded in higher-dimensional space. This is the same metric that underlies common visualisations of loss landscapes. By taking this geometric perspective literally and using the induced metric, we develop a new optimiser and compare it to existing methods, namely: SGD, Adam, AdamW, and Muon, across a range of tasks and architectures. Empirically, we conclude that this new class of optimisers is highly effective in low dimensional examples, and provides slight improvement over state-of-the-art methods for training neural networks. These new optimisers have theoretically desirable properties. In particular, the effective learning rate is automatically decreased in regions of high curvature acting as a smoothed out form of gradient clipping. Similarly, one variant of these optimisers can also be viewed as inducing an effective scheduled learning rate and decoupled weight decay is the natural choice from our geometric perspective. The basic method can be used to modify any existing preconditioning method. The new optimiser has a computational complexity comparable to that of Adam.
\end{abstract}

\newpage
\tableofcontents
\newpage

\section{Introduction}\label{sec:Intro}
The optimisation of neural networks is a cornerstone of deep learning, with the choice of optimiser often determining the success or failure of model training. Despite extensive work in this area, a fundamental disconnect persists between how practitioners visualise loss landscapes and the metrics actually employed by existing optimisation algorithms. When researchers sketch loss surfaces or generate 3D visualisations of optimisation trajectories, they implicitly impose a specific geometric structure that naturally accounts for the curvature of the loss landscape~\cite{li2018visualizing}. Yet, remarkably, this intuitive metric has never been systematically exploited in the design of practical optimisers.

This paper bridges that gap by developing a new class of optimisers based on the Riemannian metric naturally induced when the loss landscape is embedded in higher-dimensional spaces. This is precisely the metric that underlies common visualisations of loss landscapes, making our approach both principled and intuitively natural. By taking this geometric perspective literally, we derive optimisation algorithms that automatically adapt their effective learning rates based on local curvature, reducing step sizes in highly curved regions while maintaining larger updates in flatter areas.

The resulting algorithms can be viewed as a smoothed form of gradient clipping~\cite{pascanu2013difficulty}, preventing divergence when gradients become large while maintaining the benefits of a larger learning rate in flatter regions. As we will see later, one variant of the algorithm can, for appropriately chosen hyperparameters, be viewed as a form of learning rate scheduling~\cite{goyal2017accurate,loshchilov2016sgdr}. Furthermore, the decoupled form of weight-decay, as used in AdamW for example, is the more natural choice according to the geometry~\cite{loshchilov2017decoupled}. Given our claim that this metric has been the source of much of the intuition around gradient descent, the natural appearance of these methods should not be surprising. Even if the existing implementations outperform the presented approach, it is theoretically interesting to see them appear naturally from this geometric point of view. Moreover, the framework is sufficiently general that it can be applied to enhance any existing preconditioning method such as sophisticated adaptive algorithms like Muon~\cite{jordan2024muon,liu2025muon}.

From a computational perspective, our optimiser maintains the same $\mathcal{O}(N)$, where $N$ is the number of parameters, complexity as Adam, requiring only a single additional dot product computation per iteration compared to SGD. This stands in stark contrast to second-order methods, which typically demand prohibitive computational overhead, or recent innovations such as Muon, which introduce significantly higher per-iteration costs.

We validate our approach across a comprehensive suite of benchmarks, from pathological low-dimensional optimisation problems to neural network training. In low-dimensional settings, the proposed optimisers demonstrated superior performance, one such optimiser (based on log-loss embedding) being the only optimiser to successfully find the global minimum across all tested functions. These optimisers also achieved the fastest convergence times on the majority of these low dimensional problems. For neural network training, our methods prove competitive with state-of-the-art optimisers across diverse architectures and tasks, including multi-layer perceptrons (MLPs) on MNIST~\cite{mnist} and regression problems, ResNet-18~\cite{he2016deep} on CIFAR-10~\cite{krizhevsky2009learning}, and transformer~\cite{attetnion} models on language modelling tasks. On most tasks, one variant (based on RMSprop) of our custom optimisers was the best performing on average.

The purpose of this paper is twofold. First, we aim to formalise the geometric intuition behind common loss landscape visualisations, showing how standard techniques like decoupled weight decay, gradient clipping, and scheduled learning rates can naturally appear from this single perspective. Second, we apply this metric to develop new, practical optimisation algorithms. It is not our primary claim that these new optimisers will supersede all existing methods; rather, we aim to present a valuable framework for thinking about optimisation. We demonstrate that the resulting algorithms are competitive (and show slight improvement) with state-of-the-art methods for training neural networks and show a surprising level of success in low-dimensional problems. This geometric perspective also suggests several promising directions for further study, including investigating alternative embedding functions $f(\mathscr{L}(\theta))$ and developing hybrid and generalised approaches.

The field of optimisation has a history of borrowing concepts from diverse domains, with physics being a particularly rich source of inspiration. From the foundational work on simulated annealing~\cite{annealing} that connected statistical mechanics to optimisation, to recent approaches using Hamiltonian dynamics \cite{chen2014stochastic, DeLuca:2022brp} and Langevin dynamics~\cite{welling2011bayesian}. Our work continues this tradition, but through the lens of differential geometry~\cite{absil2009optimization,fei2025survey}.

We begin by introducing the optimiser and its motivation in Section~\ref{sec:Opt}; the remainder of the paper consists of benchmarks compared to existing methods. Section~\ref{sec:Path} considers finding the minimum of pathological functions, designed to disrupt gradient-based methods, typically by including narrow minima. Section~\ref{sec:Reg} considers a regression problem, where we train an MLP to approximate a randomly generated high-order polynomial of many variables. Section~\ref{sec:Class} consists of training both MLPs and ResNet-18 on the MNIST and CIFAR-10 datasets, respectively. Finally, we also consider training transformers on the TinyShakespeare~\cite{karpathy2015unreasonable} dataset.

The implementation and experiments presented here can be found in the associated GitHub repository~\cite{GH-Repo}\footnote{\url{https://github.com/harveyThomas4692/Induced-Metric-Optimiser}}. While all experiments are performed in JAX~\cite{jax2018github}, the repository also includes a PyTorch~\cite{pytorch} implementation of the optimisers.

\section{The Optimiser and Background Theory}\label{sec:Opt}
Gradient descent can be understood as a finite time-step approximation to gradient flow. Given a set of parameters, or variables, $\{\theta^i(t)|i=1\ldots N\}$ as functions of time $t$, and a loss function $\mathscr L(\theta)$, we seek solutions to the equation
\begin{equation}\label{eqn:gradFlow}
\frac{d\theta^i}{dt} = - \frac{\partial\mathscr L}{\partial \theta^i},
\end{equation}
where the stationary points at the end of the flow are typically the points of interest. Taking a finite time step $\delta t$, we can approximate a small step along this flow as
\begin{equation}\label{eqn:series}
\delta \theta^i = - \delta t\frac{\partial\mathscr L}{\partial \theta^i} + \mathcal O (\delta t^2),
\end{equation}
where the higher-order terms depend on higher derivatives of $\mathscr L$. This is the basic equation for gradient descent, and $\delta t =\eta$ is the learning rate. We will drop the higher order terms from now on.

One can also impose a metric $g_{ij}$ on the parameter space, often called gradient preconditioning in the machine learning literature~\cite{absil2009optimization,fei2025survey}. Most conventional and state-of-the-art optimisers used for training neural networks employ such a metric. In these approaches $g_{ij}$ typically depends on the training history. In other words, $g_{ij}$ is time dependent. These metrics are usually chosen to adjust the flow direction to improve numerical control. With this metric, the gradient flow, and descent, equations become
\begin{equation}\label{eqn:update}
\frac{d\theta^i}{dt} = - \sum_j g^{ij}\frac{\partial \mathscr L }{\partial\theta^j},\quad\quad
\delta\theta^i = - \eta\sum_j g^{ij}\frac{\partial \mathscr L }{\partial\theta^j},
\end{equation}
where $g^{ij}$ (with raised indices) is the matrix-inverse of $g_{ij}$.

\begin{figure}
    \centering
    \includegraphics[width=0.5\linewidth]{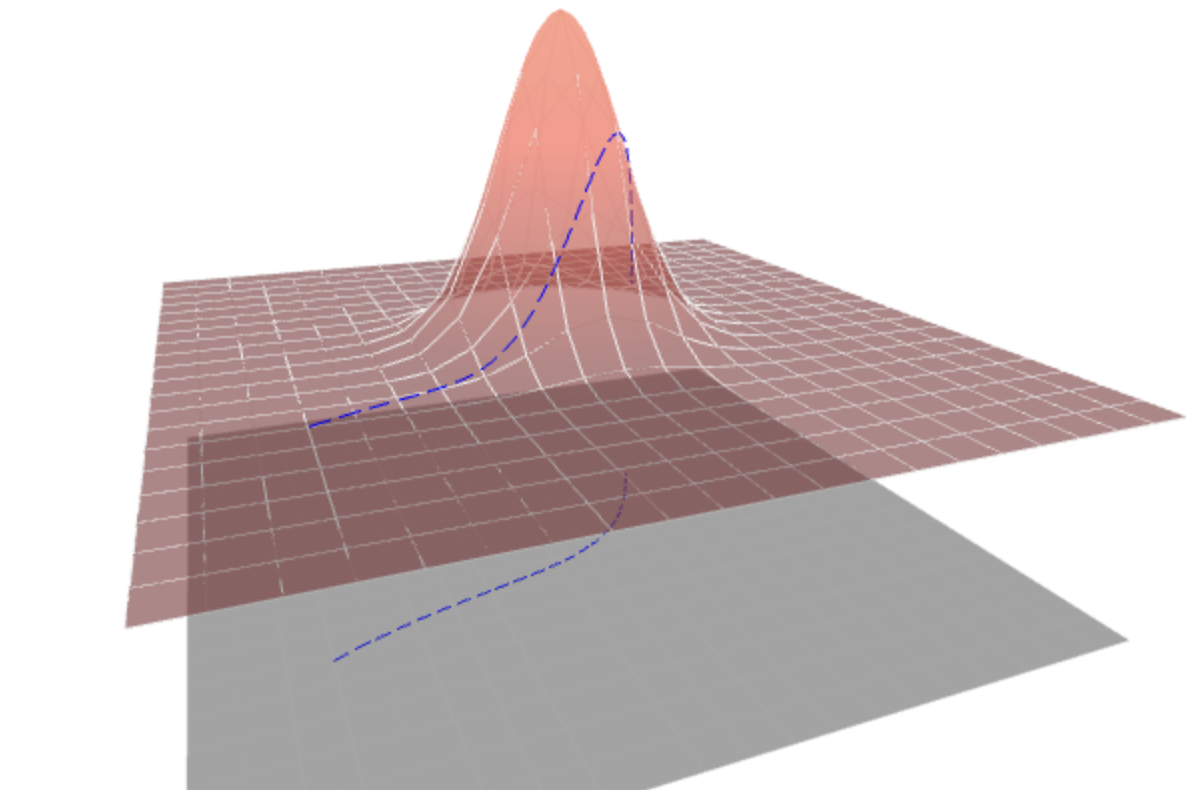}
    \caption{This plot indicates the loss landscape embedded into a larger ambient space. The euclidean metric on parameters measures distances on the grey projection, while the pull-back metric measures distances on the red surface. The blue paths indicate the trajectory on the loss landscape and its projection onto the space of parameters. In regimes of high curvature, the length measured by the pull-back metric are elongated. For the purposes of gradient descent, this has the effect of decreasing the learning rate at highly curved areas. Unlike most other preconditioning approaches, this metric does not change the direction of the trajectory away from that of steepest descent.}
    \label{fig:Projection}
\end{figure}

From a differential geometry perspective, one should think of the metric as providing a notion of distance and angle at each point in space. An infinitesimal line element at a point on a D-dimensional manifold is, in terms of coordinates ${x^i:i=1,\ldots,D}$, given by
\begin{equation}
ds^2 = \sum_{i,j}g_{ij}(x)dx^idx^j.
\end{equation}
The distance between two points is calculated by integrating this line element along the path. In differential geometry, it is crucial that under a change of coordinates, invariant quantities such as distance remain invariant. As such, the metric must also transform under a change of coordinates to respect this change. Furthermore, we need to ensure that any equations we use to describe dynamics on this curved space do not depend on the choice of coordinates—this is known as general covariance, and can be taken as the motivation for having the inverse metric in equation~\ref{eqn:update}. While we will not go into the details, we will also use general covariance as a guide through some parts of this section.

Going back to gradient descent, different choices of metrics on parameters correspond to different Riemannian manifolds. As such, different choices of metrics are, from a geometric rather than numerical perspective, somewhat arbitrary (except, perhaps, those used in natural gradient techniques~\cite{natgrad, shrestha2023natural}). However, when one visualises a loss landscape, one inevitably imposes a metric on the space. This metric is known as the pull-back metric in differential geometry. Furthermore, this metric, despite not being explicitly discussed or utilised, forms the basis for much of the intuition that presumably led to the development of various existing optimisation algorithms. The fact that this metric underpins the intuition of many practitioners, provides reason enough for further study alone. The difference between the pull-back metric on parameters and the Euclidean metric on parameters is illustrated in Figure~\ref{fig:Projection}. This metric incorporates the additional dimension, thereby increasing the distance between points in the loss landscape within regions of high curvature.

What does this metric look like? On this larger ambient space, we introduce the coordinates $\{X^M|M=1,\ldots,N+1\}$ where $X^i=\theta^i$, for $i=1,\ldots, N$, and\footnote{Note that $L$ is a coordinate on this ambient space and is not to be confused with the loss function $\mathscr L$. In other words, $L$ is simply a number that labels a position in the vertical direction of Figure~\ref{fig:Projection}.} $X^{N+1}=L$, and we introduce the metric $g_{MN}$ (we will use a bold face $\mathbf{g}$ for matrix equations) on the ambient space, given by
\begin{equation}\label{eqn:AmbMat}
    \bf{g} = \begin{pmatrix}
    {\boldsymbol\gamma} & \vec{0}  \\
    \vec{0}^T & 1
\end{pmatrix},
\end{equation}
where $\boldsymbol{\gamma}$ is some $N\times N$ matrix, that can depend on $t$, $\theta^i$ and $L$, that we need to specify. The final diagonal element is set to unity, as its value can be absorbed into the normalisation of the loss function. For the optimisers we subsequently introduce, we shall use the identity matrix and the metric implicit in RMSprop~\cite{hinton2012rmsprop}. However, as mentioned above, these choices are somewhat arbitrary and so alternative choices warrant further investigation. In particular, one could use the implicit metric from any other preconditioning algorithm, such as Muon. In any case, we shall find later that the method described here can be viewed as a type of smoothed gradient clipping.

We now pull this matrix back to our loss-landscape, given by $L=f(\mathscr{L}(\theta))$, where $f$ is some monotonic function. For geometric reasons we explain later, when we consider a non-trivial embedding function $f(\mathscr L (\theta))$ we also include this function the right hand side of gradient flow, and descent, equations
\begin{equation}
\frac{d\theta^i}{dt} = - \sum_j g^{ij}\frac{\partial}{\partial\theta^j} f(\mathscr L(\theta)),\quad\quad
\delta\theta^i = - \eta\sum_j g^{ij}\frac{\partial}{\partial\theta^j} f(\mathscr L((\theta))).
\end{equation}
However, we should emphasise that this is the author's choice, and not fundamental.

We focus on $f(\mathscr{L}(\theta)) = \mathscr L(\theta)$ for now, but extend to $f(\mathscr{L}(\theta)) = \ln(\mathscr L(\theta))$ later in this section. Given these conditions, the pull-back metric $g_{ij}$, is given by
\begin{equation}
    g_{ij} = \gamma_{ij} + \frac{\partial \mathscr L}{\partial \theta ^i}\frac{\partial \mathscr L}{\partial \theta ^j}.
\end{equation}
When the embedding is smooth, the inverse of this metric is known, via the Sherman–Morrison formula~\cite{sherman1950adjustment}, to be
\begin{equation}\label{eqn:metric}
    g^{ij} = \gamma^{ij} - \frac{\sum_{k,l}\gamma^{ik}\gamma^{lj}\frac{\partial \mathscr L}{\partial \theta ^k}\frac{\partial \mathscr L}{\partial \theta ^l}}{1+\sum_{k,l}\gamma^{kl} \frac{\partial \mathscr L}{\partial \theta ^k}\frac{\partial \mathscr L}{\partial \theta ^l}} = \gamma^{ij} - \frac{l^i  l^j}{1+\sum_{k} l_k l^k},
\end{equation}
where $\gamma^{ij}$ is the matrix inverse of $\gamma_{ij}$ and we have introduced the notation\footnote{For readers less familiar with geometry, we note that this notation is not arbitrary. There is an important distinction between upstairs and downstairs indices relating to general covariance.}
\begin{equation}\label{eqn:Notation}
    l^i = \sum_j\gamma^{ij}l_j,\quad\&\quad l_i = \frac{\partial \mathscr L}{\partial \theta ^j}.
\end{equation}
\begin{figure}[t]
    \centering
    \includegraphics[width=0.7\linewidth]{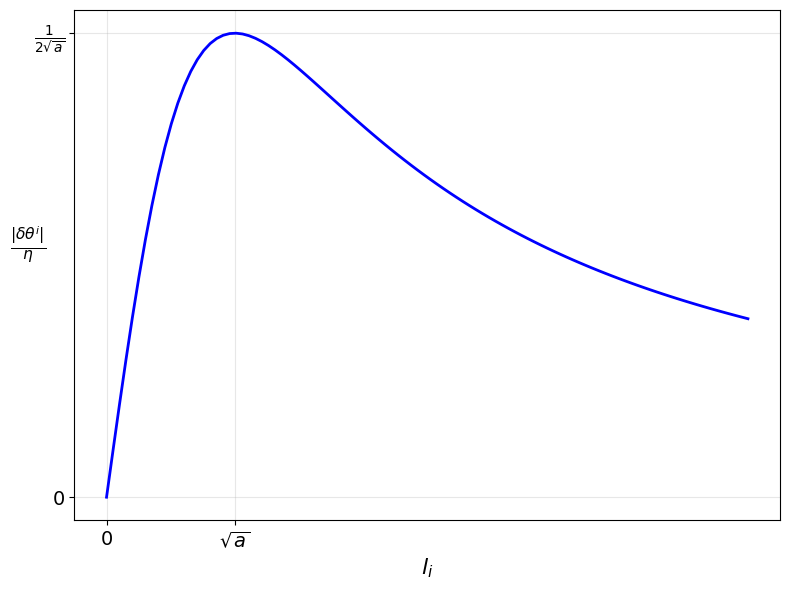}
    \caption{Gradient update profile showing the relationship between input gradient $l_i$ and parameter update $\delta\theta^i$. The curve demonstrates a smooth gradient clipping behaviour with euclidean (inverse) metric $\gamma^{ij}=\delta^{ij}$, trivial embedding function $f(\mathscr L(\theta))= \mathscr L(\theta)$, and normalisation factor $a_i=1+\sum_{k\neq i} l_k^2$ is a measure of the local curvature in the loss landscape.}
    \label{fig:grad_clip}
\end{figure}
We wish to use $g^{ij}$ to precondition our gradients, as in equation~\ref{eqn:update}. This is, given the intuition many people use in Figure~\ref{fig:Projection}, the metric that underlies all such visualisations according to geometry. A few things to note about equation~\ref{eqn:metric} with this context in mind:
\begin{itemize}
    \item \textbf{The metric is complicated, but the update is not:} $g^{ij}$ is a complicated and non-diagonal inverse metric defined on the parameters. Importantly, this metric does not depend on the training history, only on the current values of the parameters. This approach is distinct from Adam and most other preconditioning optimisers (with notable exceptions such as Muon), where the implied metric is diagonal and depends on the training process. However, upon substituting equation~\ref{eqn:metric} into equation~\ref{eqn:update}, we obtain
    \begin{equation}\label{eqn:precon}
        \begin{aligned}
            \delta \theta^i &= \eta \, l^i - \frac{\eta \, l^i  \sum_jl_jl^j}{1+\sum_{k} l_k l^k} 
            \\ &= \frac{\left(1+\sum_{k} l_k l^k\right)\eta\,l^i- \eta\,l^i\sum_jl_jl^j}{1+\sum_{k} l_k l^k}
            \\ &= -\frac{\eta\sum_j\gamma^{ij}}{1+\sum_k l_k l^k}l_j,
        \end{aligned}
    \end{equation}
    which, provided that $\gamma$ is diagonal, yields a diagonal preconditioning of the gradients.
    \item \textbf{Form of gradient clipping:} As mentioned above, equation~\ref{eqn:precon} can be viewed as a smoothed form of gradient clipping; the denominator prevents the update from diverging when the gradients are large. The form of this gradient clipping can be seen in Figure~\ref{fig:grad_clip}. 
    \item \textbf{Same computational overheads as Adam:} Using equation~\ref{eqn:precon} to precondition the gradients, with a constant $\gamma^{ij}$, has the same computation complexity as Adam. Once the gradients $l_i$ have been calculated, calculating the denominator requires a $\mathcal{O}(N)$ calculation, which is the same as calculating the first or second moments in Adam.
    \item \textbf{Can incorporate momentum and weight decay:} Momentum and weight decay can be trivially incorporated with this method. In particular weight decay can be considered as changing the gradient flow equation~\ref{eqn:update} to 
    \begin{equation}
        \frac{d\theta^i}{dt} = -\sum_j g^{ij} \frac{\partial\mathscr L}{\partial \theta^j} - \lambda \theta^i,
    \end{equation} while momentum changes this to
    \begin{equation}
        \frac{d\theta^i}{dt} = -\sum_j g^{ij} \frac{\partial\mathscr L}{\partial \theta^j} + M^i(t),
    \end{equation}
    where $M^i(t)$ is some vector field that depends on the training history\footnote{The exact form of $M^i(t)$ depends on the particular implementation of momentum.}. Alternatively, momentum can be seen as mimicking a second order differential equation. Of course both weight-decay and momentum can be considered together, and are both natural from our geometric perspective. In fact, from a differential geometry perspective, decoupled weight decay is the natural choice according to the underlying geometry due to general covariance\footnote{General covariance and weight decay could alternatively be achieved by including a $\lambda \sum_{i,j} g_{ij} \theta^i \theta^j$ term into the loss function. Since the metric required to be positive definite, this term is guaranteed to be positive. Under this formulation of weight decay decoupled and coupled weight decay become equivalent.}.
    \item \textbf{Only one new hyperparameter:} If one takes $\gamma^{ij} \propto \delta^{ij}$, then up to a redefinition of the learning rate, one has two hyperparameters to tune: $\xi$ and $\eta$, where the update is
    \begin{equation}\label{eqn:flatprecon}
       \delta \theta^i = -\frac{\eta}{1+\xi\sum_k l_k l_k}l_i.
    \end{equation}
    Since the denominator is always greater than or equal to one, the learning rate $\eta$ can likely be set significantly larger than in other algorithms. More specifically, when viewed as gradient flow, each higher-order term will also be accompanied by an inverse metric and, as such, the radius of convergence is likely to be larger. We should avoid setting $\xi$ too small, as the method would then become equivalent to traditional gradient descent, or, if $\gamma$ is non-trivial, then another preconditioning optimiser. As such, the basic version of this optimiser has only one more hyperparameter $\xi$ compared to your preferred choice of preconditioning optimiser, as the learning rate $\eta$ is usually already present in these methods. Given the form of this equation, having a sum over the $N$ paramters, a natural starting choice for $\xi$ is $\xi\sim 1/N$, where $N$ is the number of parameters. Our hyperparameter scans will be around this regime.
    \item 
    \begin{figure}[t]
        \centering
        \includegraphics[width=0.9\linewidth]{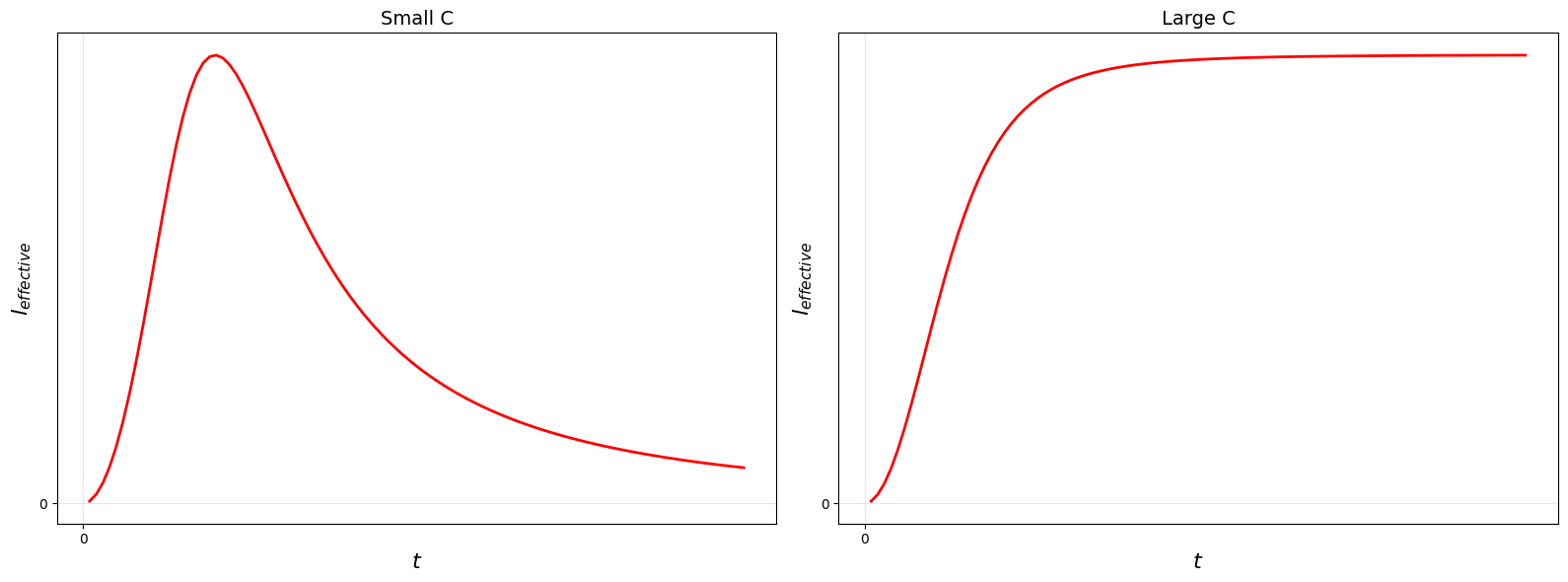}
        \caption{The effective learning rate $l_{\text{effective}}$ for the $f(\mathscr{L}(\theta)) = \log(\mathscr{L}(\theta))$ optimiser as a function of training time. For the purposes of this plot, we have assumed roughly constant gradients and $\mathscr{L} \propto t^{-p} + C$ for $p,C \in \mathbb{R}^+$. Given these assumptions, we observe that this could be considered a scheduled learning rate for appropriate choices of hyperparameters. For small $C$ compared to $t_{total}^{-p}$ we have both a learning rate warm-up and decay phases, while for $C$ large compared to $t_{total}^{-p}$ we will only observe a warm-up phase. We made these assumptions for purely pedagogical reasons, to give some intuition into the the behaviour of this optimiser during training, and should not be taken too literally. To consider the change of effective learning rate during training, some ansatz for the loss as a function of time needed to be chosen. An inverse power law was only chosen to match with the late-training scaling laws~\cite{kaplan2020scaling,hestness2017deep}. Other monotonically decreasing functions will also have a similar structure. We also assumed constant gradients to not mix this effect with the gradient clipping described earlier.}
        \label{fig:schedule}
    \end{figure}
    \textbf{Log-loss embedding function can be considered as a scheduled learning rate:} Given that the loss is always positive definite (which is frequently the case, with notable exceptions such as in reinforcement learning), we could take $f(\mathscr{L}(\theta)) = \ln(\mathscr L(\theta))$. We would then find
    \begin{equation}\label{eqn:logprecon}
       \delta \theta^i = -\frac{\eta\mathscr L(\theta)}{\mathscr L(\theta)^2+\xi\sum_k l_k l_k}l_i,
    \end{equation}
    where we see that the normalisation of the loss function cancels. Furthermore, let us make the bold, but pedagogical, assumptions that the gradients are roughly constant and that the the loss goes as $\mathscr L(\theta(t))\propto t^{-p}+C$, where $p,C\in\mathbb R^+$, during training. Given this, the rough form of the effective learning rate during training can be seen in Figure~\ref{fig:schedule}. As such, for appropriate choices of hyperparameters, this could be viewed as a a type of scheduled learning rate with both the warm-up and decay phases. The assumptions are discussed in the figure caption, but should not be taken too literally, these are used for purely pedagogical reasons.
    \item \textbf{For batched data, we can approximate the metric:} Since data is usually batched during training, we do not have an exact value for $\sum_k l_k l_k$ at each training step. This can be remedied by taking an exponential moving average (EMA), in the same manner that the first and second moments are updated in Adam. However, we expect to operate in the opposite regime with a smaller EMA parameter $\beta$. Adam typically uses $\beta$ very close to $1$, especially for the second moment. As such it prioritises a long term average of the gradients. In our approach, we require the instantaneous value of gradients but can only approximate it. This brings in a second new hyperparameter $\beta$.
    \item \textbf{The trajectories are integral curves:} This particular point is more geometrically involved than the rest of the paper, and is not crucial to the method. From a differential geometry standpoint, the gradient flow form of this optimiser (without momentum or weight decay) has a very natural interpretation. Its solutions are integral curves generated by the pullback\footnote{Where the pullback here is defined by using the metric on the loss landscape and ambient space to convert between the isomorphic 1-forms and vector fields.} of the vector field $V=-\frac{\partial}{\partial L}$ onto $L=f(\mathscr L (\theta))$. Intuitively, this is a vector field that points ``downwards", with constant magnitude, everywhere in the ambient space. This is the reason why we included $f(\mathscr L (\theta))$ on the right-hand side of the gradient flow and descent equations.
\end{itemize}
Combining the considerations above, we propose two algorithms. These are listed in Alg.~\ref{alg:custom_sgd} and Alg.~\ref{alg:custom_sgd_log}, which are based on equations~\ref{eqn:flatprecon} and~\ref{eqn:logprecon} respectively. We also include an arbitrary choice of $\gamma^{ij}$. In both cases, EMA momentum and weight decay have also been included, and the denominator is estimated via EMA. Later we also consider combining one of these with the $\boldsymbol{\gamma}^{-1}$ implied by RMSprop, and have a generalisation of AdamW as a result.

\begin{algorithm}[H]
\caption{Custom SGD with pull-back metric from loss}
\label{alg:custom_sgd}
\begin{algorithmic}[1]
\REQUIRE learning rate $\eta$, momentum coefficient $\mu$, metric coefficient $\xi$, EMA decay $\beta$, weight decay $\lambda$, inverse metric $\gamma_t^{-1}$
\REQUIRE initial parameter vector $\theta_0$
\STATE Initialise $t \leftarrow 0$, $m_0 \leftarrow \vec0$, $v_0 \leftarrow 0$
\WHILE{$\theta_t$ not converged}
\STATE $t \leftarrow t + 1$
\STATE $g_t \leftarrow \nabla_\theta f_t(\theta_{t-1})$
\STATE $\tilde{g}_t \leftarrow \gamma_t^{-1}(\cdot,g_t)$
\STATE $s_t \leftarrow \xi \cdot \sum_{i}(g^i_t\cdot \tilde{g}^i_t)$
\STATE $v_t \leftarrow \beta \cdot v_{t-1} + (1 - \beta) \cdot s_t$
\STATE $\hat{v}_t \leftarrow v_t/(1 - \beta^t)$
\STATE $r_t \leftarrow 1/(1 + |\hat{v}_t|)$
\STATE $m_t \leftarrow \mu \cdot m_{t-1} + (1-\mu)g_t$
\STATE $\hat m_t \leftarrow \gamma_t^{-1}(\cdot,m_t/(1-\mu^t))$
\STATE $\theta_t \leftarrow \theta_{t-1} - \eta \,r_t \hat{m}_t + \lambda \theta_{t-1}$
\ENDWHILE
\RETURN $\theta_t$
\end{algorithmic}
\end{algorithm}

\begin{algorithm}[H]
\caption{Custom SGD pull-back metric from log-loss}
\label{alg:custom_sgd_log}
\begin{algorithmic}[1]
\REQUIRE learning rate $\eta$, momentum coefficient $\mu$, metric coefficient $\xi$, EMA decay $\beta$, weight decay $\lambda$, inverse metric $\gamma_t^{-1}$
\REQUIRE initial parameter vector $\theta_0$
\STATE Initialise $t \leftarrow 0$, $m_0 \leftarrow \vec 0$, $v_0 \leftarrow 0$
\WHILE{$\theta_t$ not converged}
\STATE $t \leftarrow t + 1$
\STATE $g_t \leftarrow \nabla_\theta f_t(\theta_{t-1})$
\STATE $\tilde{g}_t \leftarrow \gamma_t^{-1}(\cdot,g_t)$
\STATE $L_t \leftarrow f_t(\theta_{t-1})$
\STATE $s_t \leftarrow \xi \cdot \sum_{i}(g^i_t\cdot \tilde{g}^i_t)$
\STATE $v_t \leftarrow \beta \cdot v_{t-1} + (1 - \beta) \cdot s_t$
\STATE $\hat{v}_t \leftarrow v_t/(1 - \beta^t)$
\STATE $r_t \leftarrow L_t/(L_t^2 + |\hat{v}_t|)$
\STATE $m_t \leftarrow \mu \cdot m_{t-1} + (1-\mu)g_t$
\STATE $\hat m_t \leftarrow \gamma_t^{-1}(\cdot,m_t/(1-\mu^t))$
\STATE $\theta_t \leftarrow \theta_{t-1} - \eta \, r_t\hat{m}_t + \lambda \theta_{t-1}$
\ENDWHILE
\RETURN $\theta_t$
\end{algorithmic}
\end{algorithm}

\section{Benchmarking}
The following benchmarks were all performed using JAX, Flax, and Optax. The neural networks were trained on an A100 GPU, while the low-dimensional examples were run on an M2 MacBook Pro. We used Optax's built-in optimisers for SGD with momentum~\cite{polyak1964some}, Adam, AdamW~\cite{loshchilov2017decoupled}, and the experimental version of Muon under \texttt{optax.contrib.muon}. We implemented three new optimisers in a manner that maintains compatibility with this workflow, which can be found in appendix~\ref{app:code}. The optimisers are:
\begin{itemize}
    \item \textbf{1.} Euclidean metric $\gamma_{ij}=\delta_{ij}$, and $f(\mathscr{L}(\theta)) = \mathscr L(\theta)$.
    \item \textbf{2.} Euclidean metric $\gamma_{ij}=\delta_{ij}$, and $f(\mathscr{L}(\theta)) = \ln (\mathscr L(\theta))$. We reiterate that such a choice is only possible if the loss function is always greater than zero.
    \item \textbf{3.} The metric $\gamma_{ij}(t)$ implied from RMSprop, and $f(\mathscr{L}(\theta)) = \mathscr L(\theta)$.
\end{itemize}
Note that AdamW, and SGD with EMA momentum and decoupled weight decay, exist as appropriate $\xi\rightarrow0$ limits of the third and first custom optimisers respectively, up to a redefinition of hyperparameters. Of course, many other optimisers are possible. Here, we just present the simplest version (the first optimiser), along with two variations by either changing the the embedding function $f{\mathscr (L(\theta))}$ (the second optimiser) or the metric on parameters $\gamma_{ij}(t)$ (the third optimiser).

In all cases, we do not implement a scheduled learning rate or any data augmentation. While this will likely improve the performance of all optimisers, it would probably improve them all roughly equally. As we wish to compare these optimisers on a series of fixed standard tasks, these changes are unlikely to affect the final results. We account for, and ignore, the time taken to JIT-compile each training loop.

\subsection{Examples in Low Dimensions}\label{sec:Path}
Before moving on to training neural networks, we first consider low-dimensional optimisation problems. These serve as useful examples, as there are many cases that are pathologically difficult for SGD (at least without momenta) to find the minimum, usually due to numerous local minima or highly oscillatory functions~\cite{jamil2013literature, Rosenbrock}. In particular, we consider the following functions: Ackley, Beale, Himmelblau, Rastrigin, and Rosenbrock. The details of the optimisation and following analysis can be found in the \texttt{small\_examples.ipynb} workbook in the associated GitHub repository.

For these examples, we do not consider weight decay (and so also drop AdamW). We also set $\beta$ in our custom models to unity, as the data will not be batched. A grid hyperparameter search was performed, where the hyperparameters that are shared across all optimisers (such as learning rate) spanned the same ranges across all examples. The optimisation ends when within $10^{-10}$ of the known global minimum, and the best performing run was selected for each optimiser\footnote{We define best performing as reaching the global minimum fastest. In the event that the minimum is not found, we instead select the one that got closest to the minimum.}. In all cases, the optimisers were starting from the same point away from the global minimum.

\begin{figure}[H]
    \centering
    \includegraphics[width=\linewidth]{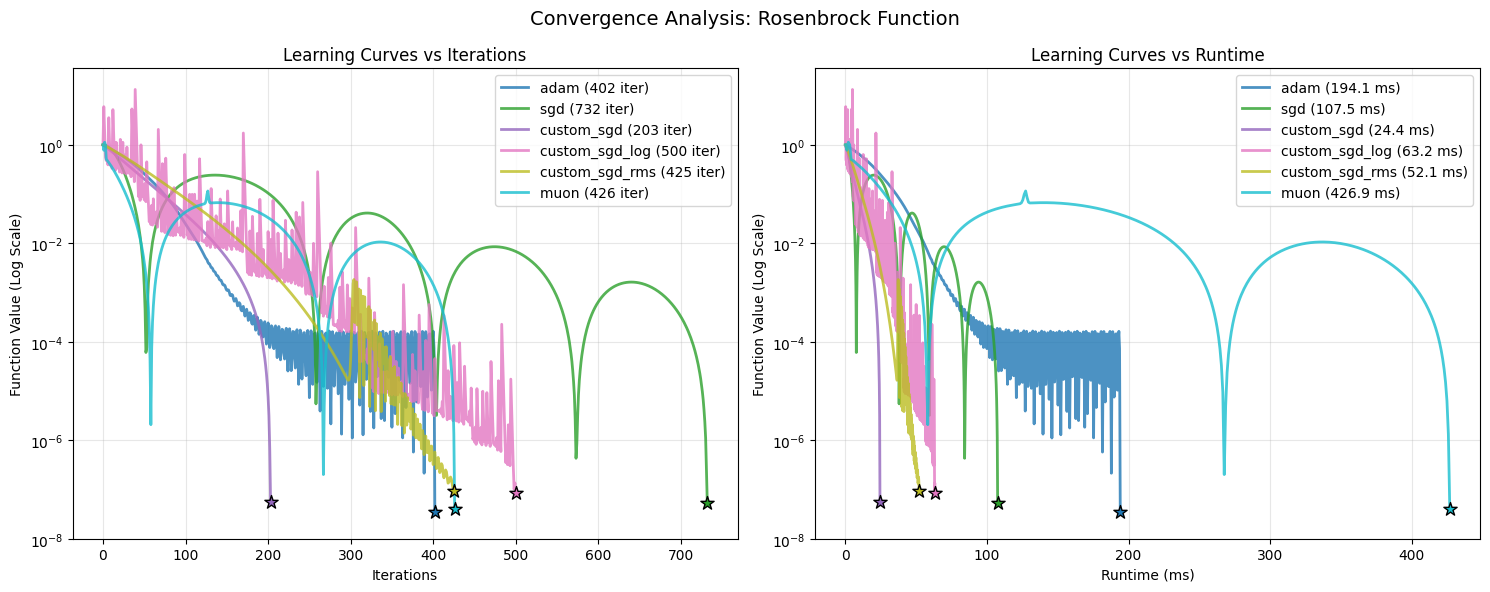}
    \caption{Convergence Analysis of the Rosenbrock function. All optimisers successfully converged. The simplest custom optimiser was the fastest in runtime and iterations.}
    \label{fig:Rosenbrock}
\end{figure}

\begin{figure}[H]
    \centering
    \includegraphics[width=\linewidth]{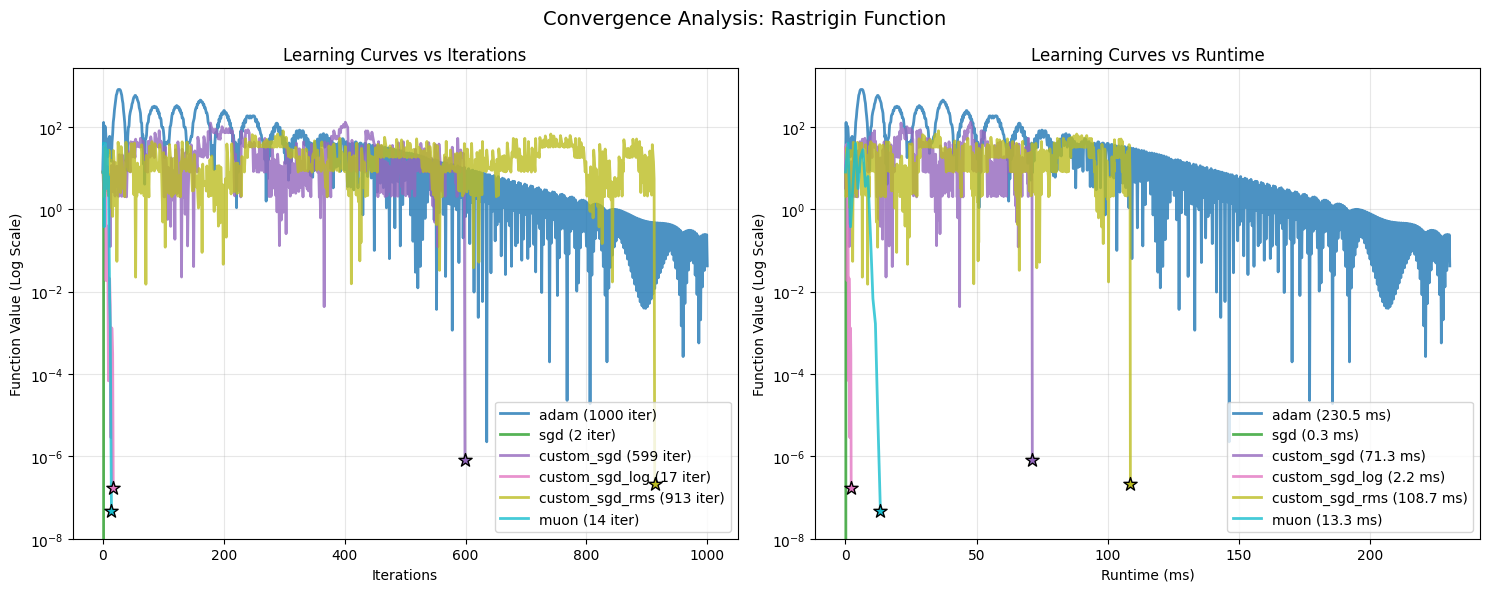}
    \caption{Convergence Analysis of the Rastrigin function. Only Adam failed to converge. SGD was the fastest both in iterations and runtime.}
    \label{fig:Rastrigin}
\end{figure}

\begin{figure}[H]
    \centering
    \includegraphics[width=\linewidth]{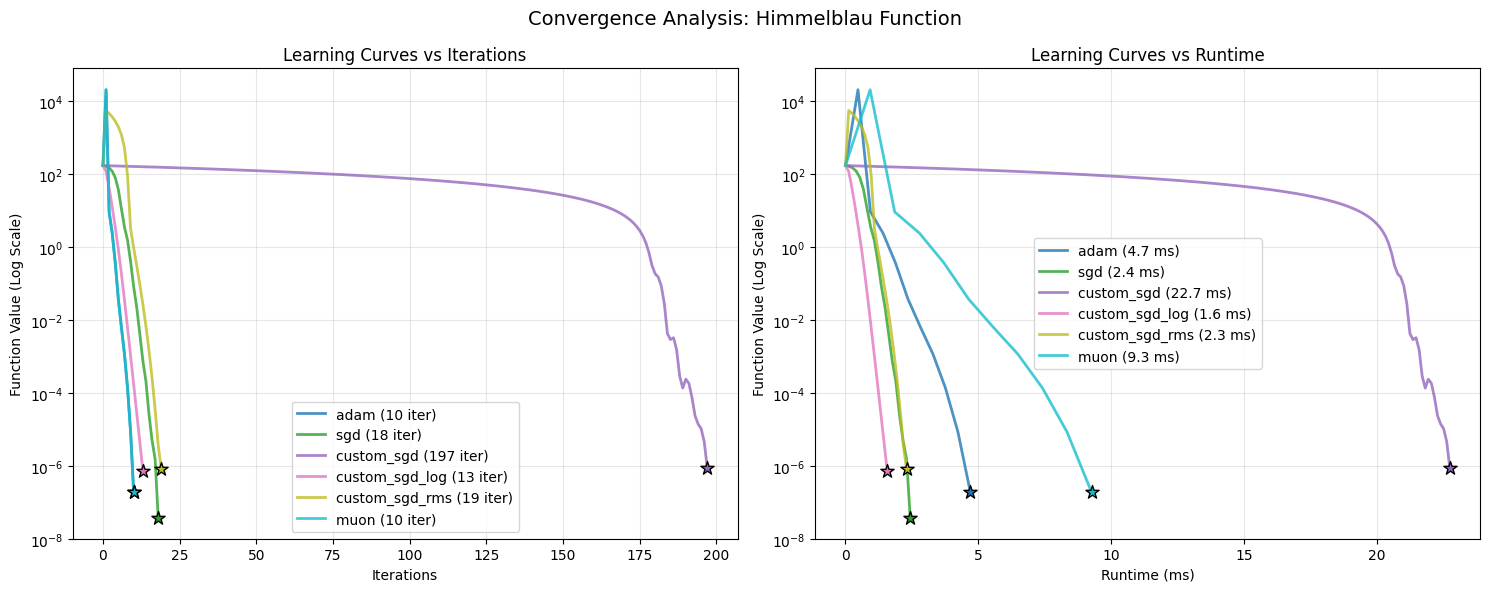}
    \caption{Convergence Analysis of the Himmelblau function. All optimisers successfully converged. Muon and Adam were joint fastest in iterations, while the log-loss custom optimiser was fastest in runtime.}
    \label{fig:Himmelblau}
\end{figure}

\begin{figure}[H]
    \centering
    \includegraphics[width=\linewidth]{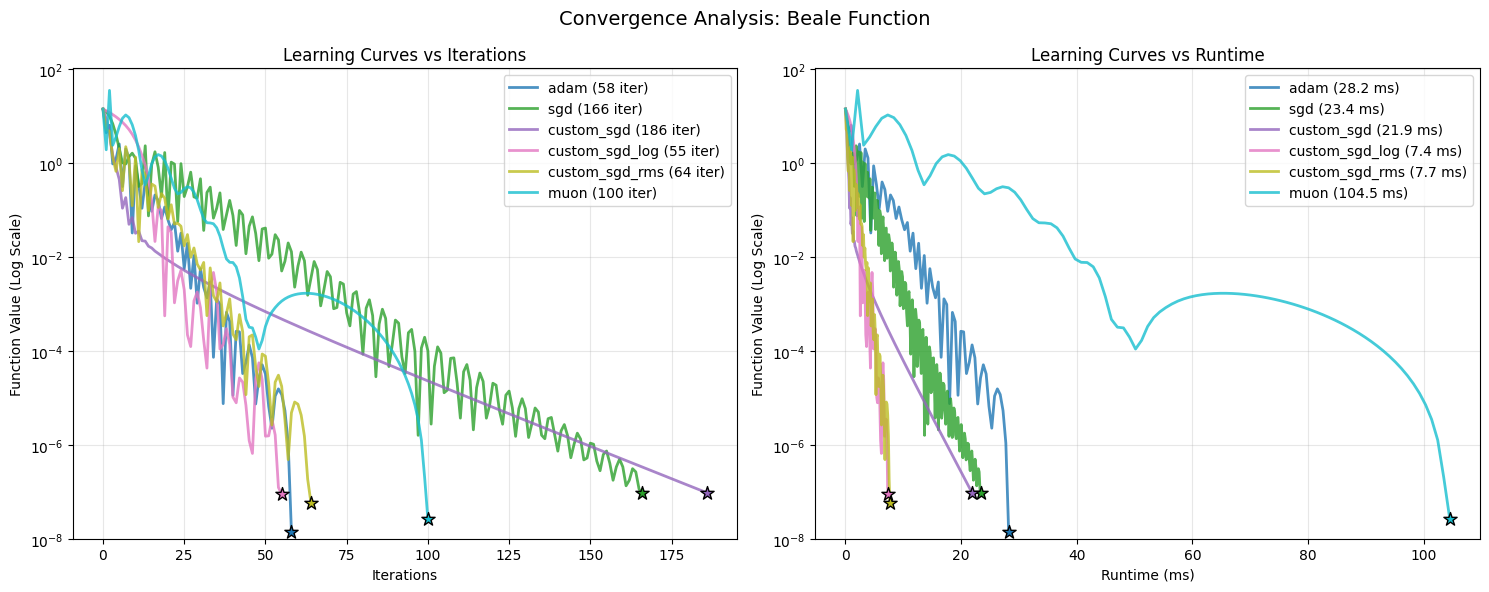}
    \caption{Convergence Analysis of the Beale function. All optimisers successfully converged. The log-loss custom optimiser was the fastest in iterations and runtime.}
    \label{fig:Beale}
\end{figure}

\begin{figure}[H]
    \centering
    \includegraphics[width=\linewidth]{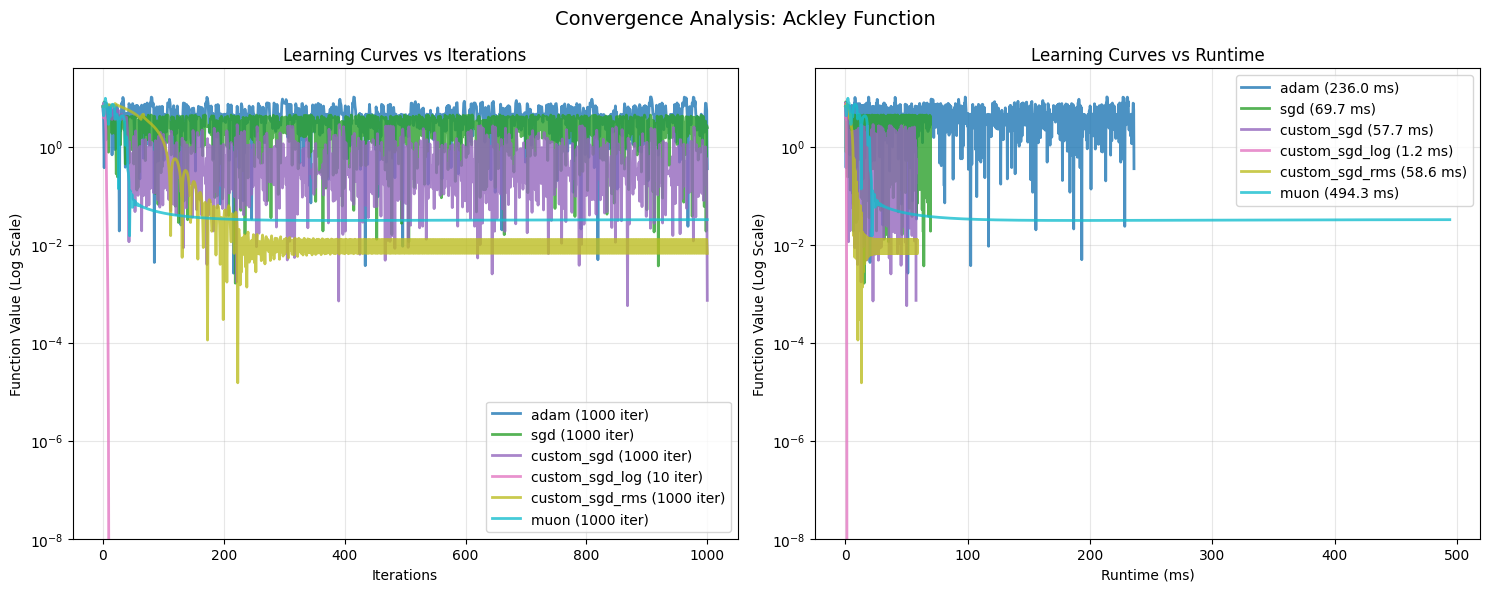}
    \caption{Convergence Analysis of the Ackley function. The log-loss custom optimiser was the only one to converge}
    \label{fig:Ackley}
\end{figure}

The results can be seen in figures~\ref{fig:Rosenbrock}, \ref{fig:Rastrigin}, \ref{fig:Himmelblau}, \ref{fig:Beale}, and \ref{fig:Ackley}. In all cases, the significantly lower computational overhead for the custom optimisers can be seen, especially compared to Muon. The log-loss optimiser is also the only optimiser that was successful in finding the minimum of all functions. The custom optimisers were typically (with the exception of the Rastrigin function) the fastest optimisers to find the minimum in runtime.

\subsection{A Regression Problem with Neural Networks}\label{sec:Reg}
\begin{table}[H]
\centering
\caption{Summary statistics for the 50 best runs with the regression task. A visualisation of the distribution can be seen in Figure~\ref{fig:regression_hist}.}
\begin{tabular}{lccccc}
\toprule
\textbf{Optimiser} & \multicolumn{2}{c}{\textbf{Min Val Loss}} & \multicolumn{2}{c}{\textbf{Min Loss Epoch}} & \textbf{Best Min Val Loss} \\
\cmidrule(lr){2-3} \cmidrule(lr){4-5}
& \textbf{Mean} & \textbf{Std} & \textbf{Mean} & \textbf{Std} & \textbf{(Epoch:Wall-Time/s)} \\
\midrule
ADAM & 0.039413 & 0.005714 & 152.6 & 47.3 & 0.021170 (184:0.21) \\
ADAMW & 0.037851 & 0.004409 & 161.3 & 37.3 & 0.025141 (196:0.24) \\
MUON & 0.043898 & 0.003694 & 155.7 & 35.0 & 0.034348 (190:0.46) \\
SGD & 0.055665 & 0.008234 & 191.0 & 11.9 & 0.036319 (196:0.20) \\
SGD\_METRIC & 0.043493 & 0.005302 & 188.2 & 20.1 & 0.031509 (198:0.20) \\
SGD\_LOG\_METRIC & 0.057149 & 0.008967 & 167.0 & 33.0 & 0.031528 (198:0.23) \\
SGD\_RMS & 0.035900 & 0.005774 & 162.1 & 42.0 & 0.022890 (178:0.22) \\
\bottomrule
\end{tabular}
\label{tab:regression}
\end{table}
In this subsection, we look at training MLPs with GELU~\cite{hendrycks2016gaussian} activations on a simple regression problem with $L_2$-loss. The data is generated from a random high degree polynomial with four variables. We keep the batch size fixed at $1024$. The sweep files and analysis workbook can be found in the GitHub repository under the names \texttt{sweep\_regression.py} and \texttt{regression\_analysis.ipynb}, respectively. In all cases, Bayesian hyperparameter optimisation, with 500 runs, were performed for all optimisers to ensure fair comparison~\cite{Bayes-Hyper}. The hyperparameters that are shared across all optimisers (such as learning rate) spanned the same ranges across all examples. 

\begin{figure}
    \centering
    \includegraphics[width=\linewidth]{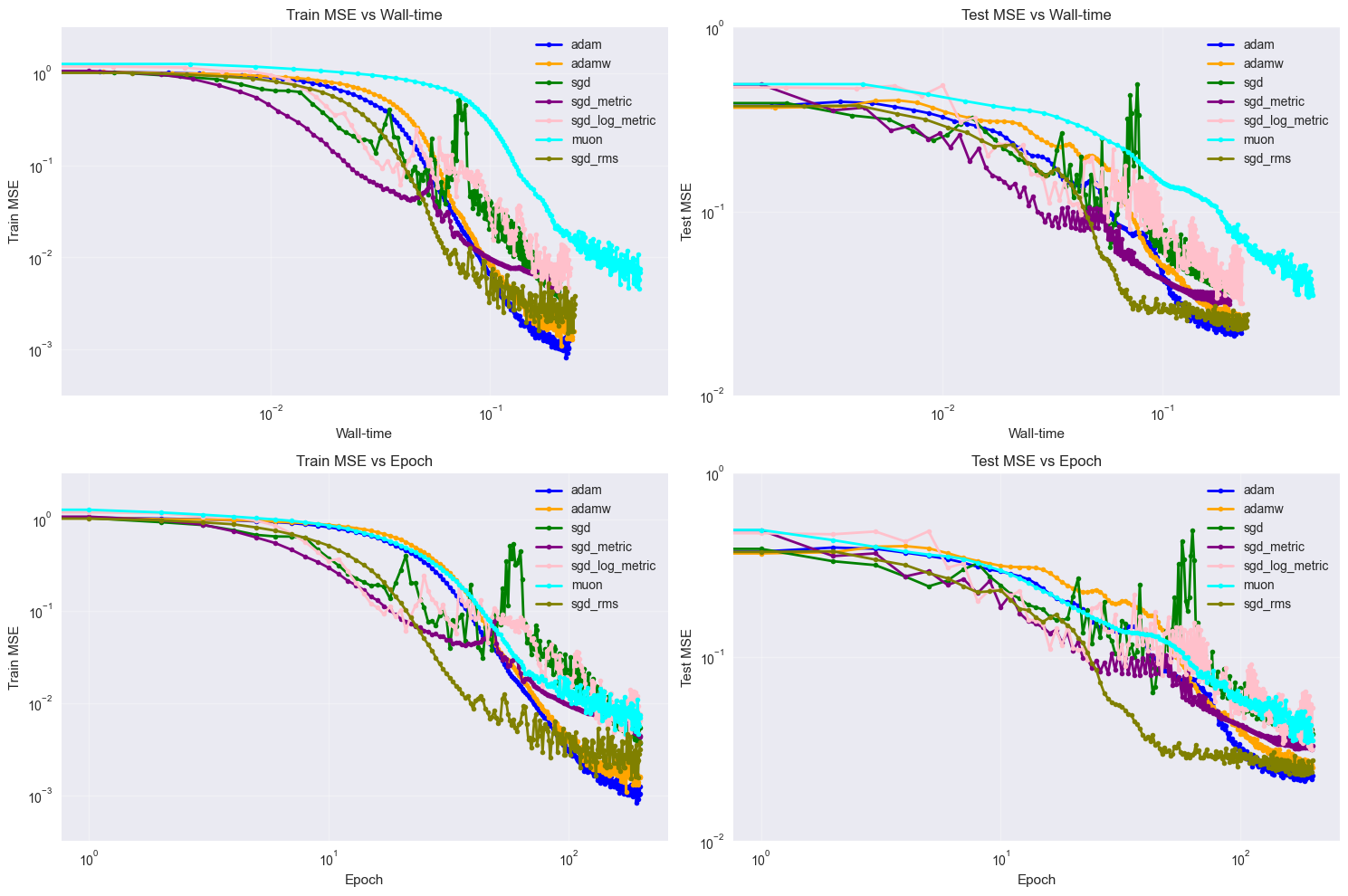}
    \caption{The training performance, on a log-log scale, for best performing run from the hyperparameter sweep for each optimiser, for a high dimensional regression problem. The final losses can be seen in Table~\ref{tab:regression}. In early training, the custom optimisers appear to outperform most of the standard methods.}
    \label{fig:regression}
\end{figure}

\begin{figure}
    \centering
    \includegraphics[width=\linewidth]{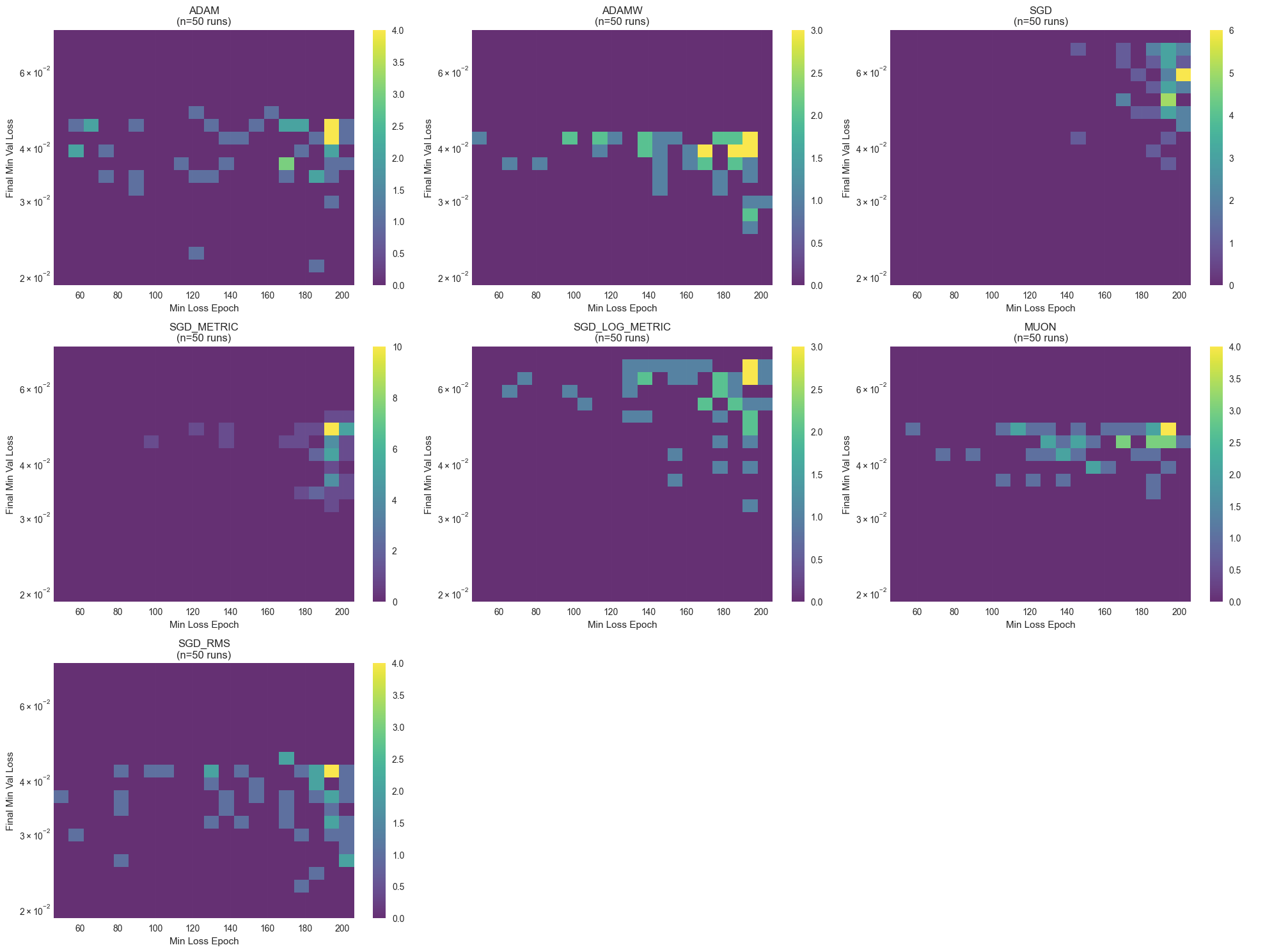}
    \caption{The best validation loss during training vs the epoch when that was achieved for each optimiser, for a high dimensional regression problem. The data is from the best 50 performing runs from the hyperparameter sweep for each optimiser. Summary statistics for these runs can be seen in Table~\ref{tab:regression}. Lower is a better model, left is faster (counted by epochs) training. The custom optimiser with the RMS metric looks like a ``spread out" version of AdamW, into regions of better performance. This is not unexpected since this optimiser contains AdamW as a special case ($\xi\rightarrow0$). Similar statements can be made about the simplest custom optimiser and SGD.}
    \label{fig:regression_hist}
\end{figure}

The training curves for each optimiser can be seen in Figure~\ref{fig:regression}, while the performance of the best fifty runs can be seen in Figure~\ref{fig:regression_hist}. Furthermore, summary statistics for the best fifty performing runs can be seen in Table~\ref{tab:regression}.

It is interesting to note that the custom optimiser with the log-loss embedding performed much worse than it did in low dimensional regression tasks. Despite Adam's best run beating both AdamW and our custom optimiser with the RMS metric, on average the best 50 runs of the latter two performed better. On average over the fifty best runs our custom optimiser with the RMS metric was the best performing optimiser.

\subsection{Classification Problems with Neural Networks}\label{sec:Class}
We first consider training an MLP with two hidden layers and GELU activations on MNIST, followed by a ResNet-18 architecture, for CIFAR-10 classification. In both cases, we keep the batch size fixed at $1024$, and shuffle the batches each epoch. The sweep files and analysis workbooks can be found in the GitHub repository under the names \texttt{sweep\_mnist\_mlp.py}, \texttt{sweep\_cifar10\_resnet18.py}, \texttt{mnist\_analysis.ipynb}, and \texttt{cifar\_analysis.ipynb}. In all cases, Bayesian hyperparameter optimisation, with 500 runs for MNIST and 200 runs for CIFAR-10, were performed for all optimisers to ensure fair comparison~\cite{Bayes-Hyper}. The hyperparameters that are shared across all optimisers (such as learning rate) spanned the same ranges across all examples. 

\begin{figure}
    \centering
    \includegraphics[width=\linewidth]{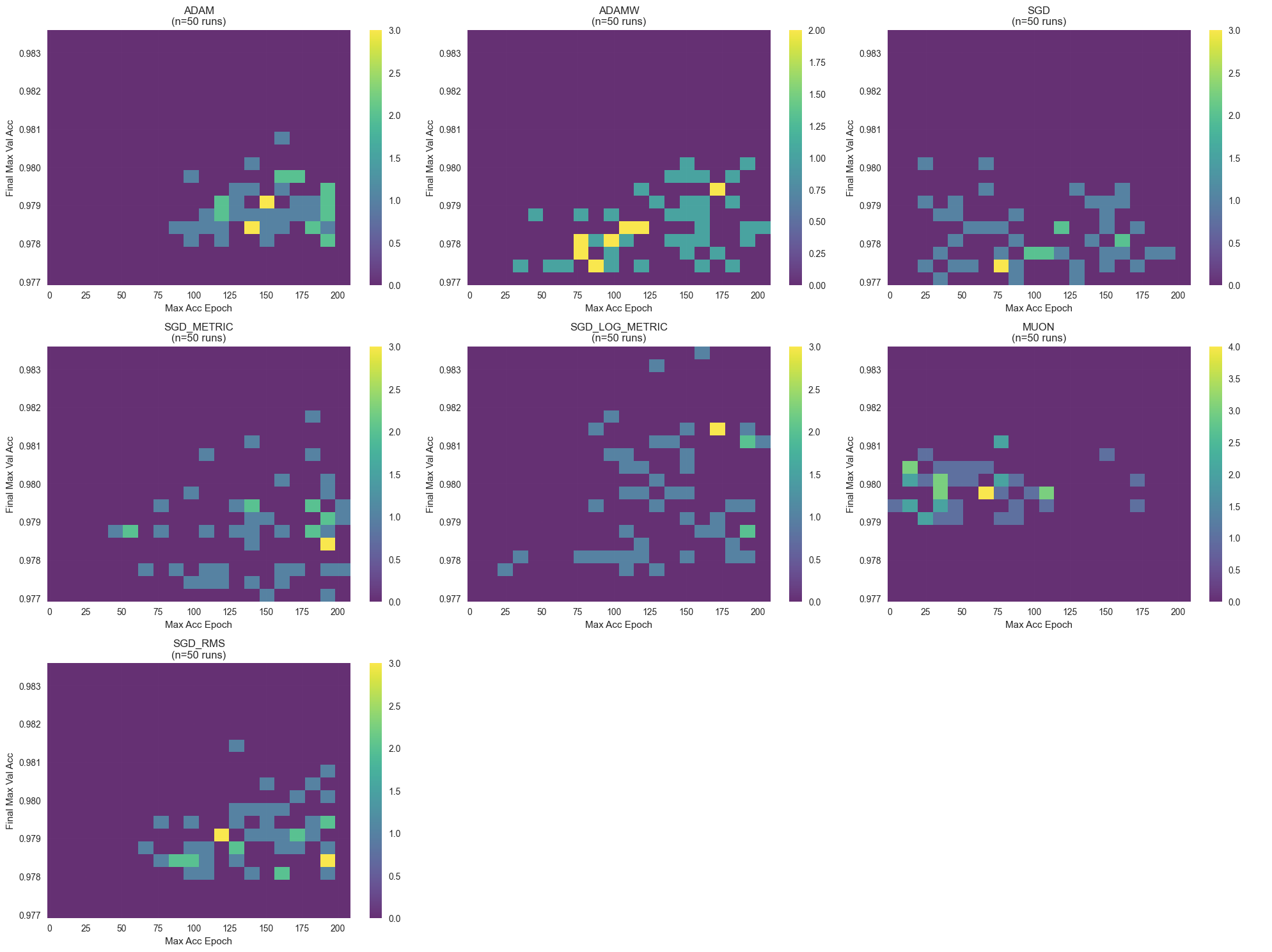}
    \caption{The best validation accuracy during training vs the epoch when that was achieved for each optimiser, for training MLPs on MNIST. The data is from the best 50 performing runs from the hyperparameter sweep for each optimiser. Summary statistics for these runs can be seen in Table~\ref{tab:mnist}. Higher is a better model, left is faster (counted by epochs) training. Once again, as with the high dimensional regression task, the custom optimiser with the RMS metric looks most similar to Adam and AdamW. In particular, it has a distribution most similar to Adam, but again ``spread out" further into faster training and higher accuracy regions. Compared to AdamW, it appears to be more accurate, but it does not train as quickly.}
    \label{fig:mnist_hist}
\end{figure}

The results from the fifty best runs on MNIST are summarised in Table~\ref{tab:mnist} and Figure~\ref{fig:mnist_hist}. All optimisation methods perform well on this task, with little variation between them. The log-loss embedding achieves a slightly higher final validation accuracy on the best run, whilst Muon performs slightly better on average. It should be noted that the speed at which SGD found its best result is somewhat misleading, as evidenced by the standard deviation in the length of the fifty best runs. Bearing in mind that the SGD result is an outlier, it remains interesting to compare how quickly each optimiser reached 98\% validation accuracy, rather than solely considering their best accuracy. For Adam, AdamW, and SGD\_RMS, this coincided with when they reached their best validation loss; however, Muon, SGD, SGD\_Metric, and SGD\_Log\_Metric achieved this milestone in 5.6s, 2.7s, 11.2s, and 4.9s, respectively. As such, the custom optimisers appear to reach models of high accuracy faster than Adam and AdamW.

\begin{table}
\centering
\caption{Summary statistics for the 50 best runs on MNIST. A visualisation of the distribution can be seen in Figure~\ref{fig:mnist_hist}.}
\begin{tabular}{lccccc}
\toprule
\textbf{Optimiser} & \multicolumn{2}{c}{\textbf{Max Val Acc}} & \multicolumn{2}{c}{\textbf{Max Acc Epoch}} & \textbf{Best Max Val Acc} \\
\cmidrule(lr){2-3} \cmidrule(lr){4-5}
& \textbf{Mean} & \textbf{Std} & \textbf{Mean} & \textbf{Std} & \textbf{(Epoch:Wall-Time/s)} \\
\midrule
ADAM & 0.978880 & 0.000567 & 151.4 & 30.4 & 0.980686 (157:16.95) \\
ADAMW & 0.978440 & 0.000772 & 128.0 & 42.6 & 0.980252 (198:21.31) \\
MUON & 0.979844 & 0.000545 & 58.5 & 39.8 & 0.981120 (81:10.48) \\
SGD & 0.978186 & 0.000744 & 102.9 & 48.9 & 0.980252 (28:2.81) \\
SGD\_METRIC & 0.978748 & 0.001023 & 146.9 & 44.2 & 0.981662 (183:18.94) \\
SGD\_LOG\_METRIC & 0.979750 & 0.001462 & 138.8 & 42.2 & 0.983290 (159:16.64) \\
SGD\_RMS & 0.979049 & 0.000763 & 143.0 & 37.0 & 0.981554 (133:14.54 ) \\
\bottomrule
\end{tabular}
\label{tab:mnist}
\end{table}

\begin{figure}
    \centering
    \includegraphics[width=\linewidth]{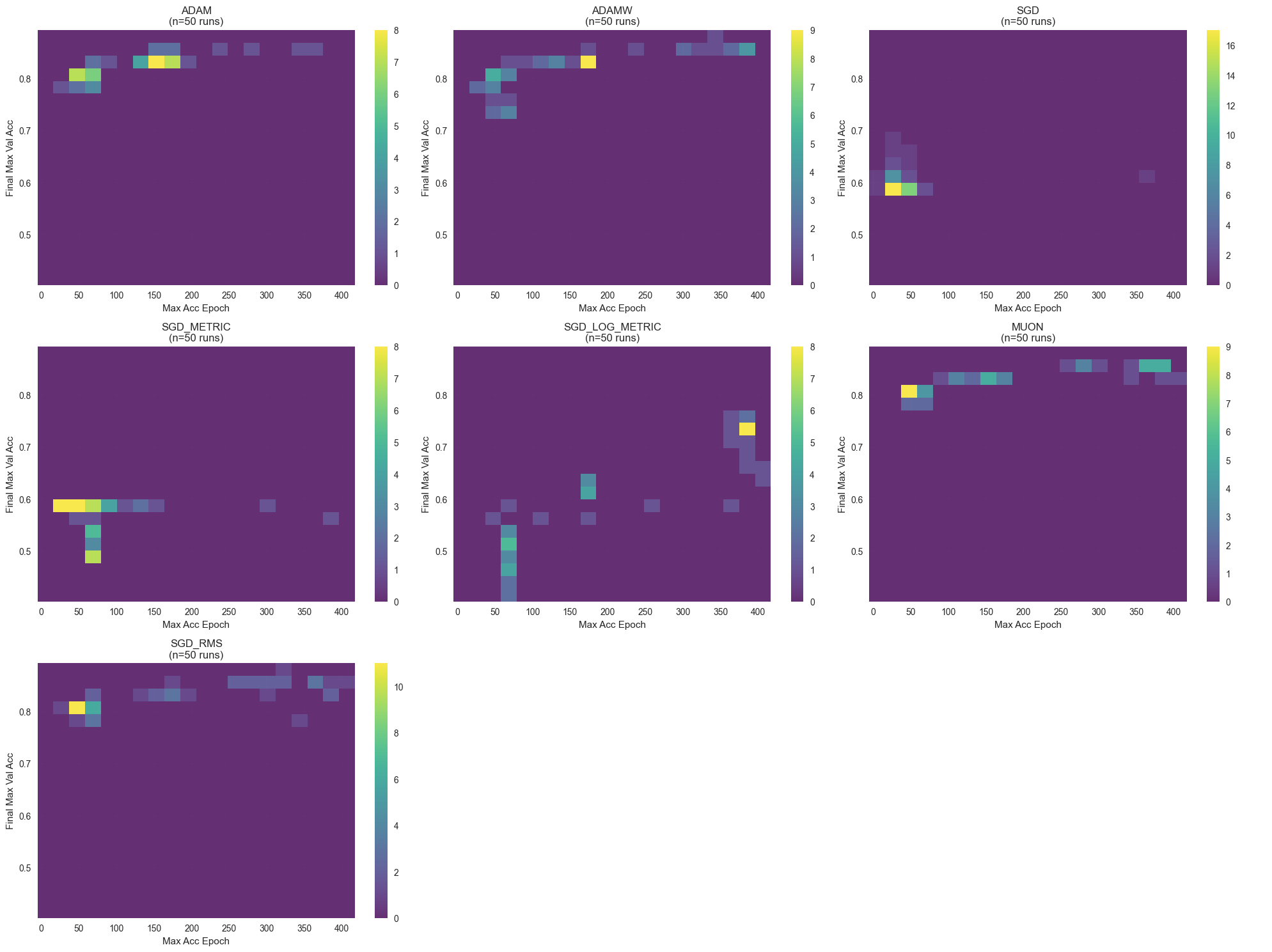}
    \caption{The best validation accuracy during training vs the epoch when that was achieved for each optimiser, for training ResNet-18 on CIFAR-10. The data is from the best 50 performing runs from the hyperparameter sweep for each optimiser. Summary statistics for these runs can be seen in Table~\ref{tab:cifar}. Higher is a better model, left is faster (counted by epochs) training. Once again, as with the high dimensional regression task, the custom optimiser with the RMS metric looks most similar to Adam and AdamW.}
    \label{fig:cifar_hist}
\end{figure}

Moving onto CIFAR-10 with ResNet-18, the best fifty runs are summarised in Table~\ref{tab:cifar} and Figure~\ref{fig:cifar_hist}. In this case, our custom optimiser with RMSprop performed best and on average achieved the highest accuracies. However, we note that for this particular case the basic version of our custom optimiser actually decreased performance when compared to SGD. Also comparing the optimiser with the log-loss embedding to SGD, we see that this did not change the average run by much, but there was a dramatic increase in the best performing runs.

\begin{table}
\centering
\caption{Summary statistics for the 50 best runs on CIFAR-10. A visualisation of the distribution can be seen in Figure~\ref{fig:cifar_hist}.}
\begin{tabular}{lccccc}
\toprule
\textbf{Optimiser} & \multicolumn{2}{c}{\textbf{Max Val Acc}} & \multicolumn{2}{c}{\textbf{Max Acc Epoch}} & \textbf{Best Max Val Acc} \\
\cmidrule(lr){2-3} \cmidrule(lr){4-5}
& \textbf{Mean} & \textbf{Std} & \textbf{Mean} & \textbf{Std} & \textbf{(Epoch:Wall-Time/s)} \\
\midrule
ADAM & 0.826033 & 0.021042 & 127.5 & 73.9 & 0.865560 (160:1519) \\
ADAMW & 0.817370 & 0.037008 & 157.6 & 117.4 & 0.869683 (349:3319) \\
MUON & 0.829572 & 0.021275 & 190.4 & 129.8 & 0.858507 (355:3348) \\
SGD & 0.600961 & 0.022746 & 43.0 & 47.7 &  0.677734 (30:286) \\
SGD\_METRIC & 0.562081 & 0.035304 & 75.2 & 63.0 & 0.594944 (45:424) \\
SGD\_LOG\_METRIC & 0.598201 & 0.106431 & 208.7 & 144.8 & 0.752387 (387:3656) \\
SGD\_RMS & 0.826309 & 0.021731 & 171.0 & 127.4 & 0.870226 (323:3094) \\
\bottomrule
\end{tabular}
\label{tab:cifar}
\end{table}

\subsection{A Language Task with Transformers}\label{sec:Tran}
We used a 4-layer GPT-style transformer (4 heads, 128-dim embeddings) for character-level language modelling on the TinyShakespeare dataset. We have use a fixed batch size of 256 throughout. The sweep files and analysis workbook can be found in the GitHub repository under the names \texttt{sweep\_shake.py} and \texttt{shake\_analysis.ipynb}, respectively. In all cases, Bayesian hyperparameter optimisation, with 500 runs, were performed for all optimisers to ensure fair comparison~\cite{Bayes-Hyper}. The hyperparameters that are shared across all optimisers (such as learning rate) spanned the same ranges across all examples. 

\begin{figure}
    \centering
    \includegraphics[width=\linewidth]{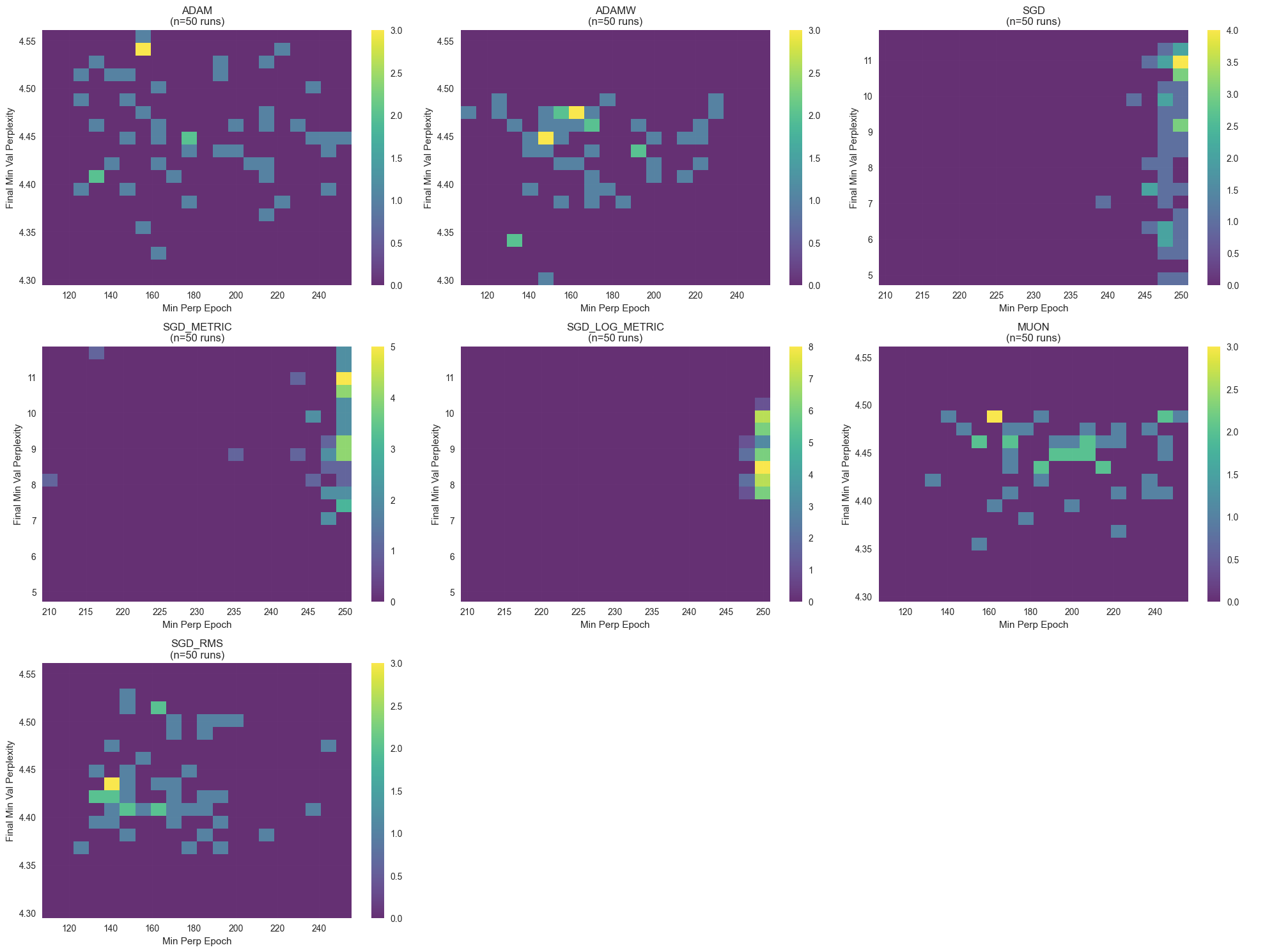}
    \caption{The best validation accuracy during training vs the epoch when that was achieved for each optimiser, for training a GPT-style transformer on TinyShakespeare. The data is from the best 50 performing runs from the hyperparameter sweep for each optimiser. Summary statistics for these runs can be seen in Table~\ref{tab:shake}. Lower is a better model, left is faster (counted by epochs) training. Note that for this plot the scales vary for the y axis, one for the better performing optimisers and the other for the worse performing ones. One should be careful comparing the three worst performing optimisers as they are far from finishing training. Once again the custom optimiser with RMSprop has a distribution similar to AdamW, but where the distribution has been shifted towards better performing and faster training.}
    \label{fig:shake_hist}
\end{figure}

The results from the fifty best runs are summarised in Table~\ref{tab:shake} and Figure~\ref{fig:shake_hist}. In this case the results split the optimisers into two distinct classes. The first class is SGD and the the first two custom optimisers, these clearly require significantly more training than the other methods and, as such, care should be taken when comparing their results. The second class of optimisers all performed well at the task. On average the custom optimiser with RMSprop performed best, though the single best run came from AdamW.

\begin{table}
\centering
\caption{Summary statistics for the 50 best runs on TinyShakespeare. A visualisation of the distribution can be seen in Figure~\ref{fig:shake_hist}.}
\begin{tabular}{lccccc}
\toprule
\textbf{Optimiser} & \multicolumn{2}{c}{\textbf{Min Val Perp}} & \multicolumn{2}{c}{\textbf{Min Perp Epoch}} & \textbf{Best Min Val Perp} \\
\cmidrule(lr){2-3} \cmidrule(lr){4-5}
& \textbf{Mean} & \textbf{Std} & \textbf{Mean} & \textbf{Std} & \textbf{(Epoch:Wall-Time/s)} \\
\midrule
ADAM & 4.455187 & 0.053800 & 180.5 & 37.1 & 4.325860 (165:8.18) \\
ADAMW & 4.437223 & 0.039655 & 168.7 & 28.8 & 4.306597 (151:6.84) \\
MUON & 4.446662 & 0.032109 & 196.3 & 31.2 & 4.355860 (155:41.01) \\
SGD & 8.643028 & 1.925947 & 247.9 & 1.6 &  5.044700 (248:8.76) \\
SGD\_METRIC & 9.364901 & 1.311992 & 246.7 & 7.2 & 7.064726 (248:8.26) \\
SGD\_LOG\_METRIC & 8.797687 & 0.753113 & 248.8 & 0.5 & 7.623210 (249:8.65) \\
SGD\_RMS & 4.432825 & 0.043723 & 164.9 & 25.8 & 4.361292 (177:8.02) \\
\bottomrule
\end{tabular}
\label{tab:shake}
\end{table}

\subsection{Summary of Experiments}\label{sec:ExpCompare}
In the previous sections we saw the custom optimisers perform extremely well for the low-dimensional pathological examples. For the training of neural networks the custom optimiser incorporating the metric from RMSprop proved to be a consistently strong performer. For the high-dimensional regression problem and the transformer-based language task, it achieved the best performance on average when compared to all other tested methods, including Adam, AdamW, and Muon. While Muon had a slightly better average on the CIFAR-10 task. These results suggest that augmenting an existing preconditioning method with the induced metric framework can lead to robust and competitive performance, and this custom optimiser appears to have a slight performance gain over Adam and AdamW.

In contrast, the log-loss embedding variant exhibited highly variable, task-dependent performance. In the low-dimensional pathological examples, it was exceptionally effective, being the only optimizer to successfully find the global minimum for every function tested. It also delivered the single best validation accuracy on the MNIST classification task. However, this success did not translate consistently to other tasks, as it performed poorly on the regression problem and the TinyShakespeare language task. Why this optimiser was successful for certain tasks only remains unclear.

\section{Conclusion}\label{sec:Conc}
This paper set out with a two-fold purpose: first, to introduce a new framework for neural network optimisation based on the commonly found visualisations of loss landscapes; and secondly, to use these ideas to design new optimisers. To this end, we proposed a new style of preconditioning optimiser, based on the induced metric on the loss landscape. This method has relatively low computational overheads, especially compared to second-order methods or Muon. Furthermore, the general method can be applied as a modification to any other preconditioning based optimiser. Many of the methods that already exist in the literature, namely: decoupled weight decay, scheduled learning rates and gradient clipping, appear naturally by taking this geometric perspective. The fact that these well-established optimisation techniques emerge naturally from our geometric perspective is noteworthy from a theoretical standpoint. These optimisers also become standard methods for appropriate choices of $\gamma^{-1}$ as $\xi\rightarrow0$. Furthermore, these methods regulate the effective learning rate in regions of high curvature, and can be viewed as a form of smoothed out gradient clipping.

These methods were shown to be highly effective for low-dimensional optimisation problems in section~\ref{sec:Path}, along with being competitive with state-of-the-art methods for the training of neural networks with various architectures, sizes, and tasks in sections~\ref{sec:Reg}, \ref{sec:Class}, and \ref{sec:Tran}. One variant of the optimiser, based on RMSprop, demonstrated slight improvement on average over Adam and AdamW.

There are several natural extensions to this work. Further comparison of this with existing methods, along with the development of hybrid methods, are desirable. In particular, ideally one would want to observe these methods when applied to significantly larger models. On the theory side, it would be interesting to explore application of geometry to optimisers further---there are many available metrics on the space of parameters and some, such as the ones presented here, may be simple to calculate and beneficial for optimisation. For example, in cases where there are $k$ independent contributions to the loss, is it beneficial to add $k$ extra dimensions, and pull-back to the loss-landscape from that space? On a similar note, is there any benefit to turning on the off-diagonal elements of the metric~\ref{eqn:AmbMat} on the ambient space, or pulling back to subsets of parameters? Lastly, are there other choices of the function $f(\mathscr L (\theta))$, that have interesting or useful properties? On this last point, given the remarkable success of the log-loss embedding in low dimensions for regression and classification, despite poorer performance on the higher dimensional tasks, it is natural to ask if there are other embedding functions that may be the reverse. The author believes that these questions warrant further investigation.
\section*{Acknowledgements}
The author would like to thank Kit Fraser-Taliente, James Halverson, Andre Lukas, Fabian Ruehle and Jesse Thaler for useful discussions. The author would also like to thank Kit Fraser-Taliente, James Halverson, Sarah Hughes and Fabian Ruehle for reading a draft version of this work.

This work is supported by the National Science Foundation under Cooperative Agreement PHY-2019786 (The NSF AI Institute for Artificial Intelligence and Fundamental Interactions, http://iaifi.org/). Some of the computations in this paper were run on
the FASRC Cannon cluster supported by the FAS Division of Science Research Computing Group at Harvard University.

\appendix
\section{Optimisers compatible with Optax}\label{app:code}
\begin{lstlisting}[caption={The three custom optimisers. These can also be found in the associated GitHub repository.}, label={lst:optimisers}]
"""
JAX implementations of custom SGD optimizers with induced metric modifications.

This module provides JAX-based optimizers using the optax framework:
- custom_sgd: Basic custom SGD with momentum and metric modification
- custom_sgd_log: Custom SGD with loss-based metric scaling
- custom_sgd_rms: Custom SGD with RMS gradient scaling

Usage examples:

# Basic custom SGD
optimizer = custom_sgd(learning_rate=0.01, momentum=0.9, xi=0.1, beta=0.1)
opt_state = optimizer.init(params)
updates, opt_state = optimizer.update(grads, opt_state, params)

# Custom SGD with loss-based metric (requires passing loss to update())
optimizer = custom_sgd_log(learning_rate=0.01, momentum=0.9, xi=0.1, beta=0.1)
opt_state = optimizer.init(params)
updates, opt_state = optimizer.update(grads, opt_state, loss, params)

# Custom SGD with RMS scaling
optimizer = custom_sgd_rms(learning_rate=0.01, momentum=0.9, xi=0.1, beta=0.1, beta_rms=0.99)
opt_state = optimizer.init(params)
updates, opt_state = optimizer.update(grads, opt_state, params)

Note: All optimizers follow the optax GradientTransformation interface for compatibility
with JAX training loops.
"""

import jax
import jax.numpy as jnp
import optax
from flax import linen as nn
from typing import NamedTuple, Optional


class SGDState(NamedTuple):
    """State container for custom SGD optimizers.
    
    Attributes:
        step: Current optimization step count
        momentum: Momentum buffer storing accumulated gradients
        metric_ema: Exponential moving average of the metric scale
        rms_ema: Optional EMA of squared gradients for RMS scaling (used in custom_sgd_rms)
    """
    step: jnp.ndarray
    momentum: jnp.ndarray  # Momentum buffer
    metric_ema: jnp.ndarray  # EMA of metric scale
    rms_ema: Optional[jnp.ndarray] = None  # EMA of RMS for custom_sgd_rms


def custom_sgd(learning_rate=0.1, momentum=0.9, xi=0.1, beta=0.8, weight_decay=0.0):
    """Custom SGD optimizer with momentum and metric modification.
    
    This optimizer implements SGD with momentum while dynamically adjusting the learning
    rate based on gradient norms. The metric modification scales updates based on an
    exponential moving average of gradient magnitudes.
    
    Args:
        learning_rate: Base learning rate (default: 0.1)
        momentum: Momentum factor for gradient accumulation (default: 0.9)
        xi: Scaling factor for gradient norm in metric computation (default: 0.1)
        beta: EMA decay rate for metric scale tracking (default: 0.8)
        weight_decay: Weight decay (L2 regularization) factor (default: 0.0)
        
    Returns:
        optax.GradientTransformation: JAX optimizer that can be used with optax
    """
    
    # Pre-compute constants for efficiency
    neg_lr = -learning_rate  # Negative for parameter updates
    one_minus_beta = 1 - beta  # For EMA computation
    one_minus_momentum = 1 - momentum  # For momentum computation

    def init(params):
        """Initialize optimizer state.
        
        Args:
            params: Model parameters to optimize
            
        Returns:
            SGDState: Initial optimizer state
        """
        return SGDState(
            step=jnp.zeros([], dtype=jnp.int32),
            momentum=jax.tree.map(jnp.zeros_like, params),
            metric_ema=jnp.zeros([]),
            rms_ema=None
        )
    
    @jax.jit
    def update(grads, state, params=None):
        """Update parameters using custom SGD with momentum and metric modification.
        
        Args:
            grads: Current gradients
            state: Current optimizer state
            params: Current parameter values (optional, used for weight decay)
            
        Returns:
            Tuple of (updates, new_state) where updates are parameter changes
        """
        step = state.step + 1
        
        # Calculate gradient norm squared across all parameters
        grad_norm_sq = optax.tree.norm(grads, ord=2, squared=True)
        trace = xi * grad_norm_sq
        
        # Update EMA of metric scale
        new_metric_ema = beta * state.metric_ema + one_minus_beta * trace
        
        # Apply bias correction and compute metric scale
        metric_corrected = new_metric_ema / (1 - beta ** step)
        metric_scale = 1 / (1 + jnp.abs(metric_corrected))
        
        # Update momentum buffer with current gradients
        new_momentum = jax.tree.map(
            lambda m, g: momentum * m + one_minus_momentum*g, 
            state.momentum, 
            grads
        )
        
        # Apply bias correction to momentum and compute parameter updates
        # Include learning rate, metric scaling, and weight decay
        updates = jax.tree.map(lambda m, p: neg_lr * metric_scale * m /(1 - momentum ** step) - learning_rate * weight_decay * p, new_momentum, params)

        # Create new optimizer state
        new_state = SGDState(step=step, momentum=new_momentum, metric_ema=new_metric_ema, rms_ema=None)
        
        return updates, new_state
    
    return optax.GradientTransformation(init, update)

def custom_sgd_log(learning_rate=0.1, momentum=0.9, xi=0.1, beta=0.8, weight_decay=0.0):
    """Custom SGD optimizer with loss-based metric modification.
    
    This optimizer extends the basic custom SGD by incorporating the loss value
    into the metric scaling computation. The metric scale is computed as a function
    of both the gradient norms and the current loss value.
    
    Args:
        learning_rate: Base learning rate (default: 0.1)
        momentum: Momentum factor for gradient accumulation (default: 0.9)
        xi: Scaling factor for gradient norm in metric computation (default: 0.1)
        beta: EMA decay rate for metric scale tracking (default: 0.8)
        weight_decay: Weight decay (L2 regularization) factor (default: 0.0)
        
    Returns:
        optax.GradientTransformation: JAX optimizer that can be used with optax
    """
    
    # Pre-compute constants for efficiency
    neg_lr = -learning_rate  # Negative for parameter updates
    one_minus_beta = 1 - beta  # For EMA computation
    one_minus_momentum = 1 - momentum  # For momentum computation

    def init(params):
        """Initialize optimizer state.
        
        Args:
            params: Model parameters to optimize
            
        Returns:
            SGDState: Initial optimizer state
        """
        return SGDState(
            step=jnp.zeros([], dtype=jnp.int32),
            momentum=jax.tree.map(jnp.zeros_like, params),
            metric_ema=jnp.zeros([]),
            rms_ema=None
        )
    
    @jax.jit
    def update(grads, state, loss, params=None):
        """Update parameters using custom SGD with loss-based metric modification.
        
        Args:
            grads: Current gradients
            state: Current optimizer state
            loss: Current loss value (required for metric computation)
            params: Current parameter values (optional, used for weight decay)
            
        Returns:
            Tuple of (updates, new_state) where updates are parameter changes
        """
        step = state.step + 1

        # Calculate gradient norm squared across all parameters
        grad_norm_sq = optax.tree.norm(grads, ord=2, squared=True)
        trace = xi * grad_norm_sq
        
        # Update EMA of metric scale
        new_metric_ema = beta * state.metric_ema + one_minus_beta * trace
        
        # Apply bias correction and compute loss-based metric scale
        metric_corrected = new_metric_ema / (1 - beta ** step)
        metric_scale = loss / (jnp.square(loss) + metric_corrected)
        
        # Update momentum buffer with current gradients
        new_momentum = jax.tree.map(
            lambda m, g: momentum * m + one_minus_momentum * g, 
            state.momentum, 
            grads
        )
        
        # Apply bias correction to momentum and compute parameter updates
        # Include learning rate, metric scaling, and weight decay
        updates = jax.tree.map(lambda m, p: neg_lr * metric_scale * m / (1 - momentum ** step) - learning_rate * weight_decay * p, new_momentum, params)

        # Create new optimizer state
        new_state = SGDState(step=step, momentum=new_momentum, metric_ema=new_metric_ema, rms_ema=None)
        
        return updates, new_state
    
    return optax.GradientTransformation(init, update)

def custom_sgd_rms(learning_rate=0.1, momentum=0.9, xi=0.1, beta=0.8, beta_rms=0.99, weight_decay=0.0, eps=1e-8):
    """Custom SGD optimizer with momentum, metric modification, and RMS scaling.
    
    This optimizer combines momentum-based SGD with adaptive gradient scaling similar
    to RMSprop. Gradients are normalized by their RMS (root mean square) values, and
    the learning rate is further modulated by a metric based on RMS-scaled gradient norms.
    
    Args:
        learning_rate: Base learning rate (default: 0.1)
        momentum: Momentum factor for gradient accumulation (default: 0.9)
        xi: Scaling factor for gradient norm in metric computation (default: 0.1)
        beta: EMA decay rate for metric scale tracking (default: 0.8)
        beta_rms: EMA decay rate for RMS computation (default: 0.99)
        weight_decay: Weight decay (L2 regularization) factor (default: 0.0)
        eps: Small constant for numerical stability (default: 1e-8)
        
    Returns:
        optax.GradientTransformation: JAX optimizer that can be used with optax
    """
    
    # Pre-compute constants for efficiency
    neg_lr = -learning_rate  # Negative for parameter updates
    one_minus_beta = 1 - beta  # For metric EMA computation
    one_minus_beta_rms = 1 - beta_rms  # For RMS EMA computation
    one_minus_momentum = 1 - momentum  # For momentum computation
    
    def init(params):
        """Initialize optimizer state.
        
        Args:
            params: Model parameters to optimize
            
        Returns:
            SGDState: Initial optimizer state with RMS tracking
        """
        return SGDState(
            step=jnp.zeros([], dtype=jnp.int32),
            momentum=jax.tree.map(jnp.zeros_like, params),
            metric_ema=jnp.zeros([]),
            rms_ema=jax.tree.map(jnp.zeros_like, params)
        )
    
    @jax.jit
    def update(grads, state, params=None):
        """Update parameters using custom SGD with momentum, metric modification, and RMS scaling.
        
        Args:
            grads: Current gradients
            state: Current optimizer state
            params: Current parameter values (optional, used for weight decay)
            
        Returns:
            Tuple of (updates, new_state) where updates are parameter changes
        """
        step = state.step + 1

        # Update EMA of squared gradients (RMS computation) for each parameter
        new_rms_ema = jax.tree.map(
            lambda r, g: beta_rms * r + one_minus_beta_rms * (g ** 2),
            state.rms_ema, grads
        )

        # Apply bias correction to RMS estimates
        rms_corrected = jax.tree.map(
            lambda r: r / (1 - beta_rms ** step),
            new_rms_ema
        )

        # Calculate RMS-scaled gradient norm for metric computation
        grad_norm_sq = jax.tree_util.tree_reduce(
        lambda acc, g_r_pair: acc + jnp.sum(g_r_pair),
        jax.tree.map(lambda g, r: g ** 2 / (jnp.sqrt(r) + eps), grads, rms_corrected),
        initializer=0.0)
        trace = xi * grad_norm_sq
        
        # Update EMA of metric scale
        new_metric_ema = beta * state.metric_ema + one_minus_beta * trace

        # Apply bias correction and compute metric scale
        metric_corrected = new_metric_ema / (1 - beta ** step)
        metric_scale = 1 / (1 + jnp.abs(metric_corrected))

        # Update momentum buffer with current gradients
        new_momentum = jax.tree.map(
            lambda m, g: momentum * m + one_minus_momentum * g, 
            state.momentum, 
            grads
        )
        
        # Apply bias correction to momentum and compute parameter updates
        # Include learning rate, metric scaling, RMS normalization, and weight decay
        updates = jax.tree.map(lambda m, p, r: neg_lr * metric_scale * m / ((1 - momentum ** step)*(jnp.sqrt(r) + eps))
                               - learning_rate * weight_decay * p, new_momentum, params, rms_corrected)

        # Create new optimizer state with updated RMS tracking
        new_state = SGDState(step=step, momentum=new_momentum, metric_ema=new_metric_ema, rms_ema=new_rms_ema)
        
        return updates, new_state
    
    return optax.GradientTransformation(init, update)
\end{lstlisting}

\bibliographystyle{ytphys}
\bibliography{sn-bibliography}

\end{document}